\documentclass[runningheads]{llncs}
\usepackage[T1]{fontenc}
\usepackage{graphicx}
\usepackage{amsmath}
\usepackage{amssymb}
\usepackage[ruled,vlined,linesnumbered]{algorithm2e}
\usepackage{booktabs}
\usepackage{makecell}
\usepackage{multirow} 
\begin{document}
\title{Subgraph Filtering for Fair Graph Neural Networks}
%
%
\author{
Haohui Lu\inst{1}\orcidID{0000-0002-5338-156X} \and
Jiyuan Tian\inst{2} \and
Fangyu Zhou\inst{3}\orcidID{0000-0001-8499-9920} \and
Shahadat Uddin\inst{3}\orcidID{0000-0003-0091-6919}
}

\authorrunning{H. Lu et al.}

\institute{
Molly Wardaguga Institute for First Nations Birth Rights, Charles Darwin University, Darwin, Australia\\
\email{haohui.lu@cdu.edu.au}
\and
Discipline of Business Analytics, The University of Sydney, Sydney, Australia\\
\email{jiyuan.tian@sydney.edu.au}
\and
School of Electrical and Computer Engineering, The University of Sydney, Sydney, Australia\\
\email{\{fangyu.zhou, shahadat.uddin\}@sydney.edu.au}
}
\maketitle              
\begin{abstract}
Graph neural networks (GNNs) can exhibit unfair behavior even when sensitive attributes are excluded from node features, because graph topology and message passing propagate group-correlated signals under sensitive homophily. Existing fairness-aware GNN methods mainly constrain representations or prediction distributions at a global level, without explicitly controlling the local structural pathways through which biased information propagates during aggregation. We propose Subgraph Filtering for Fair Graph Neural Networks (SF-GNN), a lightweight and architecture-agnostic framework that mitigates structural bias at its source. SF-GNN identifies bias-prone edges by combining sensitive homophily with structural propagation amplifiers, including hub participation and triadic closure. It then incorporates stochastic edge filtering into each message-passing step to selectively downweight or remove these edges while preserving the remaining graph structure. Training further incorporates a statistical-parity regularizer with a warm-up schedule to stabilize optimization. Experiments on five benchmark datasets show that SF-GNN achieves consistent fairness improvements while maintaining competitive predictive performance, leading to a better fairness--accuracy trade-off than recent fairness-aware GNN baselines.

\keywords{Graph neural networks \and Fairness-aware learning \and Algorithmic fairness \and Graph representation learning}
\end{abstract}
\section{Introduction}
Graph neural networks (GNNs) are increasingly deployed in high-stakes settings, including credit risk modeling, criminal justice, and social analytics \cite{zhou2020graph,yang2026relational}, where disparities across demographic groups can lead to tangible harm \cite{purificato2025gnn}. A key challenge in these domains is that graph structure itself often encodes sensitive-group correlations. In graphs characterized by sensitive homophily, message passing repeatedly aggregates and amplifies attribute-correlated signals, allowing sensitive information to influence predictions even when sensitive attributes are excluded from node features \cite{wang2024toward,sium2024individual}. This phenomenon arises from topology-driven propagation pathways and is commonly described as structural bias propagation, where graph connectivity and iterative aggregation entangle group membership with downstream predictions.

Existing fairness-aware approaches for GNNs, including adversarial debiasing, stability-based regularization, and fairness-aware encodings, have achieved important progress, but they primarily constrain representations or prediction distributions at a global level \cite{ijcai2024p50}. Consequently, they often fail to regulate the local structural pathways through which biased signals propagate during message passing, leaving topology-induced bias insufficiently addressed \cite{zhu2024fairagg}. Meanwhile, graph transformer architectures improve long-range dependency modeling through global attention, but their increased capacity to exploit graph structure can also strengthen the influence of group-correlated topology. These limitations motivate fairness mechanisms that explicitly intervene along aggregation pathways, controlling information flow through bias-prone subgraphs while preserving predictive utility \cite{wang2024toward}.

To address this gap, we introduce Subgraph Filtering for Fair Graph Neural Networks (SF-GNN), a framework that targets structural bias at its source by identifying and attenuating edges most likely to amplify sensitive-group information during aggregation. SF-GNN identifies bias-prone edges using a conjunction of sensitive homophily with structural amplifiers, including hub participation and triadic closure, and integrates stochastic edge-dependent filtering into each message-passing step. Specifically, SF-GNN selectively drops or downweights bias-prone edges while preserving the remaining graph structure, thereby weakening feedback loops that reinforce sensitive information without substantially disrupting informative connectivity. The framework is architecture-agnostic and introduces minimal computational overhead, requiring only a lightweight preprocessing stage. To stabilize optimization, the training objective further incorporates a statistical-parity regularizer with a linear warm-up schedule, reducing the risk of over-penalization during early training.

We evaluate SF-GNN on five widely used benchmark datasets and compare against standard GNNs, graph transformers, and recent fairness-aware baselines. Across datasets, SF-GNN achieves favorable fairness--accuracy trade-offs, consistently reducing statistical parity disparities while maintaining competitive predictive performance. These results highlight the effectiveness of fine-grained structural control over bias propagation in fairness-aware graph learning.

Our main contributions are summarized as follows:

\begin{enumerate}
\item We formalize a structural perspective on fairness in GNNs by identifying bias-prone edges that combine sensitive homophily with structural propagation amplifiers, including hub participation and triadic closure.

\item We propose SF-GNN, a lightweight and architecture-agnostic subgraph filtering framework that drops or downweights bias-prone edges during message passing, enabling fine-grained control over structural bias propagation without modifying the underlying graph architecture.

\item We conduct extensive experiments on benchmark datasets, including comparisons against GNN, graph transformer, and fairness-aware baselines, together with ablation and parameter sensitivity analyses, demonstrating that targeted structural filtering yields consistent fairness improvements with limited accuracy degradation.
\end{enumerate}

\section{Related Work}
\subsection{Fairness-aware GNNs}

Fairness-aware GNNs have received increasing attention as GNNs are deployed in sensitive applications where disparities across demographic groups may lead to harmful outcomes. A recurring observation is that sensitive attributes can indirectly influence predictions through graph structure and message passing, even when sensitive features are excluded from node inputs.

Standard GNN backbones such as GCN \cite{kipf2016semi} and GAT \cite{velivckovic2017graph} have demonstrated strong representation learning ability, but may amplify structural bias under sensitive homophily. GraphTrans further extends message passing through Transformer-based global attention mechanisms \cite{dwivedi2020generalization}.

Several fairness-aware GNNs have therefore been proposed. FairGNN combines sensitive attribute estimation with adversarial debiasing to reduce dependence on sensitive information \cite{dai2022learning}. NIFTY promotes counterfactual fairness and representation stability under perturbations \cite{agarwal2021towards}, while GraphAir improves fairness through automated graph augmentations \cite{ling2023learning}. BIND further mitigates unfairness through distillation-based fairness regularization \cite{dong2023interpreting}.

\subsection{Structural Fairness and Bias Propagation}

Recent work has increasingly explored structural and aggregation-aware fairness mechanisms. FairSIN reduces sensitive bias through sensitive information neutralization during message passing \cite{yang2024fairsin}, while FMP investigates fairness from a GNN architecture perspective through fairness-aware propagation mechanisms \cite{jiang2022fmp}. In graph transformers, FairGT introduces fairness-aware structural encodings and multi-hop feature integration within Transformer architectures \cite{ijcai2024p50}.

Despite these advances, many existing methods rely on adversarial optimization, graph editing, adaptive reweighting, or architectural modifications, which may introduce additional computational overhead and complexity.

\section{Preliminary}

\subsection{Problem Statement}

We consider a node classification task on an attributed graph
\[
G = (V, E, X, y, s),
\]
where $V$ is a set of $n$ nodes, $E \subseteq V \times V$ denotes edges, and
$X \in \mathbb{R}^{n \times d}$ represents node features.
Each node $v_i \in V$ is associated with a binary label $y_i \in \{0,1\}$ and
a binary sensitive attribute $s_i \in \{0,1\}$, such as gender or race.

Given a training set $V_{\text{train}} \subset V$, the goal is to learn a
GNN classifier
\[
f_\theta : (G, X) \rightarrow \hat{y},
\]
that accurately predicts node labels while reducing disparities between
different sensitive groups.

Unlike conventional learning settings with independent samples, GNNs update node representations through message passing, which aggregates information from neighboring nodes \cite{kipf2016semi}. As a result, correlations between sensitive attributes and graph connectivity can be repeatedly reinforced during aggregation, allowing sensitive information to indirectly influence predictions even when $s$ is excluded from input features. This motivates fairness-aware graph learning methods that explicitly consider structural bias propagation.

\subsection{Fairness Concepts in GNNs}

Let $h_i^{(l)}$ denote the representation of node $v_i$ at layer $l$ in a
graph neural network.
A general message passing layer can be expressed as
\[
h_i^{(l+1)} =
\phi \!\left(
h_i^{(l)},
\bigoplus_{j \in \mathcal{N}(i)} \psi(h_i^{(l)}, h_j^{(l)}, e_{ij})
\right),
\]
where $\mathcal{N}(i)$ is the neighborhood of node $i$,
$\psi(\cdot)$ is a message function, $\oplus$ denotes an aggregation
operator, and $\phi(\cdot)$ is an update function.

In many real-world graphs, sensitive attributes exhibit strong correlation
with graph connectivity.
Formally, for a sensitive attribute $s \in \{0,1\}$, homophily implies
\[
\Pr(s_i = s_j \mid (i,j) \in E) \gg \Pr(s_i \neq s_j),
\]
resulting in neighborhoods dominated by nodes from the same sensitive group.
Through repeated message passing, such correlations can be amplified across
layers, causing learned representations $h_i^{(l)}$ to encode
sensitive-group information even when sensitive attributes are excluded
from the input features.

From a fairness perspective, this phenomenon can be viewed as
structural bias propagation, where biased correlations embedded in
graph topology are accumulated through iterative aggregation.
As a result, disparities in prediction outcomes may arise from the
interaction between graph structure and message passing dynamics, rather
than solely from feature-level bias.

These observations motivate fairness-aware graph learning approaches that
consider not only model objectives and representations, but also how
information propagates through graph structure during message passing.
In particular, regulating the influence of bias-prone structural patterns
offers a potential pathway to mitigate unfairness while preserving
predictive performance.

\subsection{Evaluation Metrics}

We evaluate both predictive performance and group fairness using accuracy,
statistical parity difference, and equality of opportunity.

\paragraph{Accuracy (ACC)}
Accuracy measures the proportion of correctly classified nodes:
\[
\mathrm{ACC} = \frac{1}{|V|} \sum_{i \in V} \mathbb{I}\big(\hat{y}_i = y_i\big),
\]
where $V$ denotes the set of evaluated nodes, $y_i \in \{0,1\}$ is the
ground-truth label of node $i$, $\hat{y}_i$ is the predicted label, and
$\mathbb{I}(\cdot)$ denotes the indicator function, which equals 1 if its
argument is true and 0 otherwise.

\paragraph{Statistical Parity Difference (SPD)}
Statistical parity difference quantifies group-level disparity by measuring
the difference in positive prediction rates across sensitive groups:
\[
\Delta\mathrm{SP} =
\left|
\Pr(\hat{y}=1 \mid s=0) - \Pr(\hat{y}=1 \mid s=1)
\right|,
\]
where $s_i \in \{0,1\}$ denotes the sensitive attribute of node $i$.
Lower values of $\Delta\mathrm{SP}$ indicate fairer outcomes, with
$\Delta\mathrm{SP}=0$ indicating perfect statistical parity.

\paragraph{Equality of Opportunity Difference (EO)}
Equality of opportunity evaluates whether true positive rates are similar
across sensitive groups.
We measure the equality of opportunity difference as
\[
\Delta\mathrm{EO} =
\left|
\Pr(\hat{y}=1 \mid y=1, s=0)
-
\Pr(\hat{y}=1 \mid y=1, s=1)
\right|.
\]
Lower values of $\Delta\mathrm{EO}$ indicate fairer treatment of positive
instances across groups, while $\Delta\mathrm{EO}=0$ corresponds to equal
true positive rates.

Higher ACC reflects stronger predictive performance, while lower
$\Delta\mathrm{SP}$ and $\Delta\mathrm{EO}$ indicate reduced dependence of
predictions on sensitive-group membership.
In this work, we seek to improve fairness by reducing both
$\Delta\mathrm{SP}$ and $\Delta\mathrm{EO}$ without substantially
sacrificing ACC.

\section{Method}

We present Subgraph Filtering for Fair Graph Neural Networks (SF-GNN), a fairness-aware graph learning framework that regulates bias propagation during message passing through subgraph filtering. Rather than treating all homophilic edges as equally problematic, SF-GNN focuses on homophilic edges that are also structurally positioned to amplify sensitive-group information.

\subsection{Overview}

\begin{figure*}[h]
  \centering
  \includegraphics[width=\textwidth]{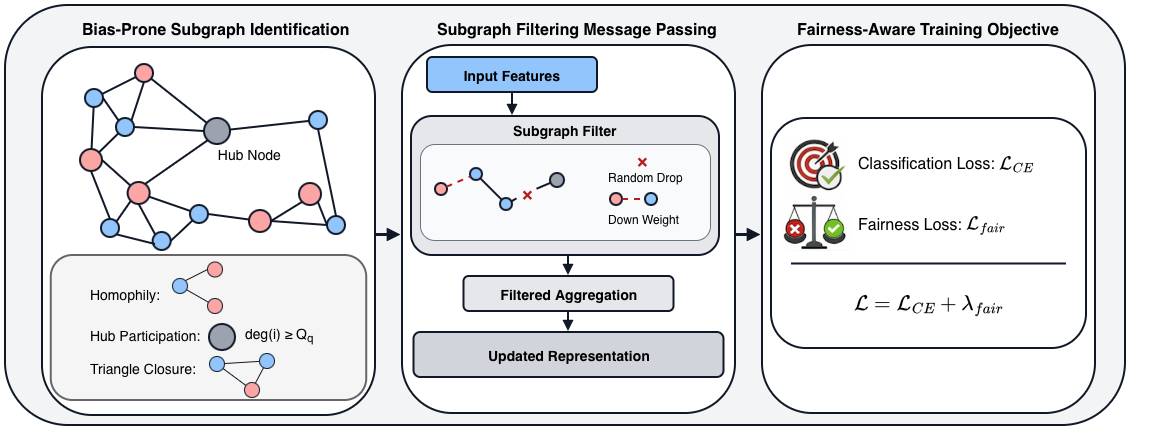}
  \caption{Overview of the SF-GNN framework.}
  \label{fig:sfgenn}
\end{figure*}

SF-GNN operates in three stages. Figure~\ref{fig:sfgenn} shows the framework overview. First, we identify bias-prone edges using sensitive homophily together with two structural propagation amplifiers: hub participation and triadic closure. Second, we integrate a stochastic subgraph filtering mechanism into message passing, where bias-prone edges are selectively dropped or downweighted during aggregation. Third, model training is guided by a fairness-aware objective that balances predictive accuracy and group fairness. This design complements feature-level and objective-level fairness methods by directly intervening in structural pathways through which sensitive-group information may propagate.

\subsection{Bias-Prone Subgraph Identification}

Let $G=(V,E)$ be an undirected graph, where $V$ denotes the set of nodes and $E$ denotes the set of edges. Let $s_i \in \{0,1\}$ denote the sensitive attribute of node $v_i \in V$. An edge $(i,j)\in E$ is considered homophilic if $s_i=s_j$. While sensitive homophily can indicate potential bias transmission, homophily alone does not fully determine whether an edge will strongly affect message passing. We therefore combine homophily with structural propagation amplifiers.

Ignoring nonlinearities, a $L$-layer GNN propagates information approximately as
\[
Z^{(L)} \approx \tilde{A}^{L} X W,
\]
where
\[
\tilde{A}=D^{-1/2}(A+I)D^{-1/2}
\]
denotes the normalized adjacency matrix with self-loops, $X$ is the node-feature matrix, and $W$ denotes learnable transformations. Under this formulation, the influence of node $j$ on node $i$ after $L$ propagation layers depends on the number and strength of propagation paths between them.

\paragraph{Homophily.}
Sensitive homophily indicates whether an edge may carry group-correlated information:
\[
H_{ij}=\mathbb{I}(s_i=s_j),
\]
where $\mathbb{I}(\cdot)$ is the indicator function.

\paragraph{Hub Participation.}
High-degree nodes participate in many aggregation operations and may spread correlated information to many neighborhoods. Let $\mathrm{deg}(i)$ denote the degree of node $i$, and let $Q_q$ denote the $q$-th quantile of the degree distribution. An edge is associated with a hub if
\[
U_{ij}=
\mathbb{I}\big(
\mathrm{deg}(i)\ge Q_q
\lor
\mathrm{deg}(j)\ge Q_q
\big).
\]

\paragraph{Triadic Closure.}
Edges involved in triangles belong to tightly connected local communities, where repeated aggregation may reinforce homophilic feedback loops. We define
\[
T_{ij}=
\mathbb{I}\big(
|\mathcal{N}(i)\cap\mathcal{N}(j)|>0
\big).
\]

\paragraph{Bias-Prone Edge Mask.}
SF-GNN identifies a bias-prone edge as
\[
B_{ij}=
H_{ij}
\land
(U_{ij}\lor T_{ij}),
\]
where $B_{ij}=1$ indicates that the edge is both sensitive-homophilic and structurally positioned to amplify group-correlated information during repeated message passing. This criterion does not assume that hubs or triangles are universal carriers of bias. Rather, it prioritizes sensitive-homophilic edges that are structurally more likely to repeatedly propagate correlated signals across neighborhoods. The resulting binary mask $B$ can be computed once before training and reused across epochs.

\subsection{Subgraph Filtering Message Passing}

Let $h_i^{(l)} \in \mathbb{R}^d$ denote the representation of node $i$ at layer $l$, and let $\mathcal{N}(i)$ denote its neighborhood. A standard message-passing update can be written as
\[
h_i^{(l+1)}
=
\phi\left(
h_i^{(l)}
+
\sum_{j \in \mathcal{N}(i)}
m_{ij}^{(l)}
\right),
\quad
m_{ij}^{(l)}
=
\psi(h_j^{(l)}),
\]
where $\psi(\cdot)$ is a message function and $\phi(\cdot)$ is an update function.

SF-GNN modifies this process by introducing an edge-dependent filtering weight $\alpha_{ij}$:
\[
h_i^{(l+1)}
=
\phi\left(
h_i^{(l)}
+
\sum_{j \in \mathcal{N}(i)}
\alpha_{ij}\,
\psi(h_j^{(l)})
\right).
\]

For bias-prone edges $(i,j)$ with $B_{ij}=1$, the filtering weight $\alpha_{ij}$ is sampled as
\[
\alpha_{ij} =
\begin{cases}
0,
& \text{with probability } p_{\text{drop}}, \\

\gamma,
& \text{with probability } 1-p_{\text{drop}},
\end{cases}
\qquad
0 < \gamma < 1.
\]

For non-bias-prone edges with $B_{ij}=0$, we set
\[
\alpha_{ij}=1.
\]

This stochastic filtering mechanism attenuates propagation pathways while preserving the remaining graph structure. Residual connections and normalization layers are used to maintain expressive capacity and training stability.
\subsection{Model Architecture}

SF-GNN is designed as a lightweight filtering module that can be incorporated into message-passing GNN backbones. In our main implementation, we use two subgraph-filtering convolution layers followed by a multilayer perceptron classifier. Each convolution layer consists of a linear transformation, subgraph-filtered aggregation, residual connection, batch normalization, and nonlinearity.

\subsection{Fairness-Aware Training Objective}

Model training minimizes the composite objective
\[
\mathcal{L}
=
\mathcal{L}_{\text{CE}}
+
\lambda_{\text{fair}}
\mathcal{L}_{\text{fair}},
\]
where $\mathcal{L}_{\text{CE}}$ denotes the cross-entropy loss over labeled nodes.

We adopt statistical parity as the fairness regularizer and optimize a differentiable surrogate using predicted probabilities. Let $\hat{p}_i$ denote the predicted probability of the positive class for node $i$. The fairness loss is defined as
\[
\mathcal{L}_{\text{fair}}
=
\left|
\mathbb{E}_{i:s_i=0}[\hat{p}_i]
-
\mathbb{E}_{i:s_i=1}[\hat{p}_i]
\right|.
\]

To avoid over-penalizing fairness early in training, $\lambda_{\text{fair}}$ is increased linearly from $0$ to its target value during a warm-up period. Sensitive attributes are used only during training to compute the bias-prone edge mask and fairness regularizer, and are not required during inference.

\subsection{Computational Complexity}

Let $n=|V|$ denote the number of nodes, $m=|E|$ the number of edges, $d$ the hidden dimension, and $L$ the number of GNN layers. Degree computation requires $\mathcal{O}(m)$ time, while triangle detection via adjacency-set intersection costs $\mathcal{O}\!\left(\sum_{(i,j)\in E}\min(\mathrm{deg}(i),\mathrm{deg}(j))\right)$. This preprocessing is performed once before training. During message passing, SF-GNN performs weighted edge aggregation with complexity $\mathcal{O}(md)$ per layer, matching standard message-passing GNNs up to constant-time edge operations. The overall training complexity is $\mathcal{O}(Lmd + T_{\text{prep}})$, where $T_{\text{prep}}$ denotes the preprocessing cost of bias-prone edge detection. Memory complexity is $\mathcal{O}(nd+m)$.

\section{Experiments}

In this section, we evaluate the effectiveness of SF-GNN in improving group
fairness while preserving predictive performance.\footnote{The code for SF-GNN is available at \url{https://github.com/haohuilu/SF-GNN}.}

\subsection{Datasets}

We evaluate SF-GNN on five widely used fairness-aware graph learning benchmarks: NBA, German, Credit, Income, and Pokec\_n. Across all datasets, node classification is treated as the downstream task, and each dataset contains a designated sensitive attribute that should not unduly influence predictions.

NBA \cite{dai2021say} contains professional basketball players from the 2016--2017 season, where nodes represent players and edges capture social relationships derived from Twitter interactions. The dataset contains 403 nodes and 16,570 edges. Nationality (U.S.\ vs.\ non-U.S.) is treated as the sensitive attribute, and the task is to predict whether a player's salary exceeds the median.

German \cite{asuncion2007uci} is a credit-risk graph constructed from bank client records, containing 1,000 nodes and 22,242 edges. Gender is treated as the sensitive attribute, and the task is to classify clients as low-risk or high-risk credit holders.

Credit \cite{yeh2009comparisons} models credit card users, where nodes represent individuals and edges encode similarities in spending and repayment behavior. The dataset contains 30,000 nodes and 137,377 edges. Age is treated as the sensitive attribute, and the task is to predict default on credit card payments.

Income \cite{asuncion2007uci} is derived from the Adult Census dataset and contains 14,821 nodes and 100,483 edges. Edges are constructed based on feature similarity following prior work \cite{agarwal2021towards}. Race is treated as the sensitive attribute, and the task is to predict whether annual income exceeds \$50{,}000.

Pokec\_n \cite{takac2012data} is a social network dataset derived from a Slovak online social platform. Nodes represent users and edges represent friendship relations. Following prior fairness-aware graph learning studies, we use region information as the sensitive attribute and predict users' working-field labels.

For datasets with more than two classes, we binarize labels by mapping classes $\{0,1\}$ to $1$ and all remaining classes to $0$, following common practice in fairness-aware graph learning. We randomly split nodes into 50\% training, 25\% validation, and 25\% test sets following established baselines \cite{agarwal2021towards,ijcai2024p50,dong2023interpreting}.

\subsection{Experimental setup}
All experiments are implemented in PyTorch with PyTorch Geometric utilities \cite{fey2019fast}.
We use a two-layer GCN architecture with a hidden
dimension of 256, followed by a multilayer perceptron classifier.
Models are trained using the Adam optimizer with an initial learning rate of
0.005 and weight decay set to $10^{-4}$.
Training is performed for up to 500 epochs, with a cosine annealing learning
rate schedule applied throughout optimization.
To mitigate class imbalance, we apply balanced class weights in the
cross-entropy loss.

For SF-GNN, the fairness regularization coefficient $\lambda$, edge drop probability $p$, 
downweight factor $\gamma$, and hub quantile $q$ are tuned via grid search over 
$\lambda \in \{0.5, 0.6, 0.7, 0.8\}$, 
$p \in \{0.1, 0.2, 0.3\}$, 
$\gamma \in \{0.7, 0.8, 0.9\}$, 
and $q \in \{0.4, 0.5, 0.6\}$.

The fairness loss is linearly warmed up over the first 100 epochs to avoid
over-regularization at early training stages.
Bias-prone edges are identified once before training by default, and the mask
is reused across epochs to reduce computational overhead.

All experiments are lightweight enough to run on a CPU. For transparency and reproducibility, all hyperparameter configurations and evaluation metrics are logged during training.

\subsection{Baselines}
To ensure a comprehensive and fair comparison, we evaluate SF-GNN against a
diverse set of baseline methods spanning three categories:
\emph{standard graph neural networks (GNNs)}, \emph{graph transformers (GTs)},
and \emph{fairness-aware GNNs}.
This design allows us to assess performance trade-offs across expressive
power, structural modeling capacity, and fairness interventions.

\paragraph{Standard GNNs.}
We include widely used message-passing GNN architectures as performance
baselines.
Specifically, we evaluate GCN \cite{kipf2016semi} and
GAT \cite{velivckovic2017graph}, which represent representative convolutional
and attention-based GNNs.

\paragraph{Graph Transformers.}
To evaluate the impact of global attention mechanisms, we include
GraphTrans \cite{wu2021representing}, which employs Transformer-based
self-attention to capture long-range node interactions.

\paragraph{Fairness-aware GNNs.}
We compare against several representative fairness-aware graph learning
methods that explicitly aim to mitigate bias in node classification.
FairGNN \cite{dai2022learning} reduces dependence on sensitive
attributes through sensitive attribute estimation and adversarial debiasing.
NIFTY \cite{agarwal2021towards} promotes counterfactual fairness and
stability under sensitive-attribute perturbations.
BIND \cite{dong2023interpreting} addresses bias through distillation-based
regularization, while GraphAir \cite{ling2023learning} improves fairness
through fairness-aware graph augmentations.
We further include recent fairness-aware methods including
FairSIN \cite{yang2024fairsin},
FMP \cite{jiang2022fmp},
and FairGT \cite{ijcai2024p50}, which focus on sensitive information
neutralization, fairness-aware message passing, and fairness-aware graph
transformers, respectively.

Across all baselines, we follow the original implementations and recommended
hyperparameter settings whenever possible, and apply the same data splits and
evaluation metrics to ensure fair comparison.

\subsection{Results}
Table~\ref{tab:main_results} reports the performance of all methods across five benchmark datasets. Overall, SF-GNN achieves a strong balance between predictive performance and group fairness. Compared with standard GNNs, graph transformers, and fairness-aware baselines, SF-GNN reduces fairness disparities while maintaining competitive or improved classification accuracy.

On NBA and Pokec\_n, SF-GNN achieves the highest ACC among all compared methods while substantially reducing $\Delta$SP relative to standard GNNs and graph transformers. In particular, on NBA, SF-GNN improves ACC from 71.65 (FAIRGT) to 72.15 while reducing $\Delta$SP from 11.16 to 2.29. On Pokec\_n, SF-GNN achieves the best overall ACC (91.31) together with the lowest $\Delta$SP (1.11), demonstrating strong effectiveness on large-scale socially homophilic graphs.

On German and Credit, SF-GNN also remains highly competitive. On German, SF-GNN achieves the highest ACC (70.56) while maintaining substantially lower fairness disparity than most standard GNN and graph transformer baselines. On Credit, SF-GNN achieves the second-highest ACC (72.51), close to FAIRGT (72.94), while preserving low $\Delta$SP and competitive $\Delta$EO. Although some fairness-aware baselines such as FairGNN and NIFTY obtain slightly lower $\Delta$SP on Credit, they often do so at the expense of predictive performance or stability across datasets.

On Income, SF-GNN attains an ACC of 74.58 with $\Delta$SP of 3.38, offering a balanced trade-off relative to GraphTrans, which achieves higher ACC (77.45) but substantially larger fairness disparity ($\Delta$SP of 22.49), and FairGNN, which reports the lowest $\Delta$SP (2.65) at the cost of markedly reduced accuracy (45.56).

Compared with fairness-aware baselines, SF-GNN generally achieves higher predictive performance while maintaining comparable or improved fairness metrics. These results suggest that directly regulating structural bias propagation during message passing can be more effective than relying solely on adversarial debiasing, representation regularization, graph augmentation, or global fairness constraints.

Across all datasets, SF-GNN demonstrates stable performance under different graph structures and sensitive attributes, supporting the effectiveness of subgraph filtering as a lightweight and architecture-agnostic mechanism for mitigating topology-induced bias in graph neural networks.

\begin{table*}[t]
\centering
\caption{Performance comparison across five datasets. ACC ($\uparrow$) indicates classification accuracy, while lower $\Delta$SP ($\downarrow$) and $\Delta$EO ($\downarrow$) indicate fairer outcomes. Results are averaged over 10 runs. Best results are shown in \textbf{bold}.}
\label{tab:main_results}
\setlength{\tabcolsep}{1pt}
\resizebox{\textwidth}{!}{
\begin{tabular}{lccc ccc ccc ccc ccc}
\toprule
\textbf{Method}
& \multicolumn{3}{c}{\textbf{NBA}}
& \multicolumn{3}{c}{\textbf{German}}
& \multicolumn{3}{c}{\textbf{Credit}}
& \multicolumn{3}{c}{\textbf{Income}}
& \multicolumn{3}{c}{\textbf{Pokec\_n}} \\

\cmidrule(lr){2-4}
\cmidrule(lr){5-7}
\cmidrule(lr){8-10}
\cmidrule(lr){11-13}
\cmidrule(lr){14-16}

& ACC$\uparrow$ & $\Delta$SP$\downarrow$ & $\Delta$EO$\downarrow$
& ACC$\uparrow$ & $\Delta$SP$\downarrow$ & $\Delta$EO$\downarrow$
& ACC$\uparrow$ & $\Delta$SP$\downarrow$ & $\Delta$EO$\downarrow$
& ACC$\uparrow$ & $\Delta$SP$\downarrow$ & $\Delta$EO$\downarrow$
& ACC$\uparrow$ & $\Delta$SP$\downarrow$ & $\Delta$EO$\downarrow$ \\
\midrule

GCN
& 62.53 & 9.50 & 13.86
& 65.12 & 6.45 & 4.22
& 66.03 & 3.45 & 6.42
& 63.83 & 14.68 & 13.00
& 84.91 & 1.56 & 2.82 \\

GAT
& 65.57 & 10.43 & 11.00
& 65.12 & 7.78 & 7.62
& 67.29 & 3.47 & 7.06
& 62.46 & 16.34 & 13.29
& 70.75 & 4.45 & 4.44 \\

GraphTrans
& 69.87 & 13.42 & 20.30
& 70.08 & 7.29 & 6.15
& 70.10 & 10.27 & 15.51
& \textbf{77.45} & 22.49 & 12.18
& 89.67 & 1.29 & 2.17 \\

\midrule

FairGNN
& 63.29 & 10.44 & \textbf{8.89}
& 61.12 & 6.67 & 10.47
& 69.86 & 0.81 & 1.87
& 45.56 & \textbf{2.65} & \textbf{1.92}
& 85.15 & 1.69 & 2.07 \\

NIFTY
& 68.61 & 11.02 & 17.37
& 64.56 & 4.95 & \textbf{2.47}
& 66.91 & \textbf{0.69} & 2.41
& 58.81 & 4.24 & 3.89
& 81.39 & 2.34 & 2.22 \\

BIND
& 66.08 & 7.83 & 11.04
& 63.44 & 5.21 & 5.63
& 56.61 & 3.20 & 6.24
& 64.47 & 14.26 & 12.40
& 84.79 & 1.99 & 1.96 \\

GraphAir
& 68.10 & 9.26 & 13.93
& 67.44 & 5.17 & 4.52
& 71.55 & 1.00 & 2.73
& 76.03 & 4.13 & 6.26
& 87.53 & 2.39 & 3.03 \\

FairSIN
& 64.81 & 11.07 & 9.08
& 63.04 & 7.36 & 10.28
& 67.78 & 0.83 & \textbf{1.71}
& 59.24 & 6.55 & 4.75
& 85.30 & 1.59 & \textbf{1.40} \\

FMP
& 66.33 & 7.68 & 9.60
& 62.48 & 6.58 & 10.42
& 71.84 & 1.18 & 2.28
& 72.37 & 3.51 & 5.06
& 87.14 & 2.75 & 2.57 \\

FairGT
& 71.65 & 11.16 & 23.22
& 69.52 & \textbf{1.80} & 3.04
& \textbf{72.94} & 0.84 & 2.93
& 75.75 & 3.04 & 3.54
& 88.87 & 2.11 & 4.44 \\

\midrule

\textbf{SF-GNN}
& \makecell{\textbf{72.15}\\$\pm$4.00}
& \makecell{\textbf{2.29}\\$\pm$2.16}
& \makecell{9.10\\$\pm$7.94}

& \makecell{\textbf{70.56}\\$\pm$2.81}
& \makecell{3.12\\$\pm$3.11}
& \makecell{6.44\\$\pm$4.48}

& \makecell{72.51\\$\pm$0.99}
& \makecell{1.48\\$\pm$1.27}
& \makecell{2.81\\$\pm$1.64}

& \makecell{74.58\\$\pm$1.63}
& \makecell{3.38\\$\pm$1.46}
& \makecell{3.06\\$\pm$1.26}

& \makecell{\textbf{91.31}\\$\pm$0.15}
& \makecell{\textbf{1.11}\\$\pm$0.12}
& \makecell{3.26\\$\pm$2.67} \\

\bottomrule
\end{tabular}
}
\end{table*}

\subsection{Ablation Study}
Table~\ref{tab:ablation_full} presents the ablation study of SF-GNN across five benchmark datasets. We keep the same backbone and training protocol while modifying one component at a time to isolate its contribution. Specifically, we compare the full model against variants without fairness regularization (\textit{No fairness}), no structure filtering (\textit{No SF}), alternative filtering strategies (\textit{Drop}, \textit{Downweight}), different bias-prone edge definitions (\textit{HO}, \textit{HH}, \textit{HT}), and training/control variants (\textit{No FWU}, \textit{RBPM}, \textit{RFSNE}). Overall, the full SF-GNN achieves the best or near-best balance between predictive performance and fairness across datasets. Removing fairness regularization consistently increases $\Delta$SP and $\Delta$EO, particularly on Income and Credit, indicating that fairness control is essential for mitigating bias. Meanwhile, variants without structured filtering or fairness warm-up generally produce weaker fairness results, supporting the importance of depth-aware fairness scheduling and stable optimization. These results demonstrate that the combination of progressive fairness control and structural filtering contributes substantially to the fairness–utility trade-off achieved by the proposed model.

\begin{table*}[h]
\centering
\caption{Ablation study of Subgraph Filtering Graph Neural Network (SF-GNN) components across datasets.
SF-GNN denotes the full proposed model.
No fairness removes fairness regularization.
No SF removes structure filtering.
Drop and Downweight apply only edge dropping or edge downweighting.
HO uses homophily-only edge identification,
HH uses homophily + hub participation,
and HT uses homophily + triangle closure.
No FWU disables fairness warm-up,
RBPM recomputes the bias-prone mask each epoch,
and RFSNE randomly filters the same number of edges.
We report ACC ($\uparrow$), $\Delta$SP ($\downarrow$), and $\Delta$EO ($\downarrow$). Best results for each metric are highlighted in bold.}
\label{tab:ablation_full}
\resizebox{\linewidth}{!}{
\begin{tabular}{ll
|cccc
|cccc
|ccc}
\toprule
\multirow{2}{*}{\textbf{Dataset}} 
& \multirow{2}{*}{\textbf{Metric}} 
& \multicolumn{4}{c|}{\textbf{Filtering Mechanism}} 
& \multicolumn{4}{c|}{\textbf{Bias-Prone Edge Definition}} 
& \multicolumn{3}{c}{\textbf{Training / Control}} \\
\cmidrule(lr){3-6}
\cmidrule(lr){7-10}
\cmidrule(lr){11-13}
& 
& SF-GNN 
& No fairness
& No SF 
& Drop
& Downweight
& HO
& HH
& HT
& No FWU
& RBPM
& RFSNE \\
\midrule

\multirow{3}{*}{NBA}
& ACC$\uparrow$
& \textbf{72.15}
& 70.38
& 72.13
& 70.67
& 72.06
& 72.05
& 72.01
& 71.05
& 72.06
& 70.13
& 72.01 \\

& $\Delta$SP$\downarrow$
& \textbf{2.29}
& 6.54
& 9.20
& 8.63
& 6.61
& 4.44
& 8.80
& 3.84
& 14.05
& 5.89
& 7.43 \\

& $\Delta$EO$\downarrow$
& \textbf{9.10}
& 14.02
& 19.44
& 10.76
& 11.81
& 14.18
& 13.63
& 14.22
& 11.30
& 19.17
& 9.17 \\
\midrule

\multirow{3}{*}{German}
& ACC$\uparrow$
& 70.56
& \textbf{71.28}
& 70.24
& 70.32
& 70.08
& 70.32
& 69.52
& 70.24
& 70.24
& 69.60
& 70.40 \\

& $\Delta$SP$\downarrow$
& 3.12
& 4.47
& 1.74
& 1.74
& \textbf{0.68}
& 1.60
& 2.07
& 2.30
& 1.71
& 1.80
& 2.12 \\

& $\Delta$EO$\downarrow$
& 6.44
& \textbf{2.61}
& 4.62
& 3.35
& 3.45
& 3.32
& 3.98
& 4.00
& 4.16
& 3.39
& 4.22 \\
\midrule

\multirow{3}{*}{Credit}
& ACC$\uparrow$
& 72.51
& 71.25
& 71.24
& 71.44
& 72.27
& 72.27
& 71.81
& 72.32
& \textbf{73.42}
& 72.31
& 72.02 \\

& $\Delta$SP$\downarrow$
& 1.48
& 8.91
& 1.06
& 1.03
& 1.08
& \textbf{0.61}
& 1.20
& 1.15
& 1.00
& 1.19
& 1.18 \\

& $\Delta$EO$\downarrow$
& 2.81
& 17.43
& \textbf{1.68}
& 2.03
& 3.02
& 4.09
& 3.09
& 2.78
& 3.48
& 3.04
& 2.80 \\
\midrule

\multirow{3}{*}{Income}
& ACC$\uparrow$
& 74.58
& \textbf{76.31}
& 73.64
& 73.98
& 74.33
& 74.12
& 73.86
& 73.98
& 74.39
& 73.94
& 74.38 \\

& $\Delta$SP$\downarrow$
& 3.38
& 21.84
& 2.88
& 2.84
& 4.10
& \textbf{2.60}
& 3.71
& 3.12
& 3.55
& 2.88
& 3.25 \\

& $\Delta$EO$\downarrow$
& 3.06
& 11.83
& 4.19
& 2.86
& 2.28
& 2.35
& 2.42
& 2.13
& 2.44
& \textbf{1.62}
& 4.18 \\
\midrule

\multirow{3}{*}{Pokec\_n}
& ACC$\uparrow$
& \textbf{91.31}
& 91.45
& 91.04
& 90.66
& 91.66
& 91.07
& 91.56
& 91.20
& 91.09
& 91.19
& 91.29 \\

& $\Delta$SP$\downarrow$
& 1.11
& \textbf{0.57}
& 1.42
& 0.99
& 1.16
& 1.19
& 1.42
& 1.11
& 1.28
& 1.14
& 1.06 \\

& $\Delta$EO$\downarrow$
& 3.26
& \textbf{1.32}
& 4.88
& 3.66
& 5.18
& 4.95
& 5.41
& 4.27
& 5.62
& 4.29
& 3.94 \\
\bottomrule
\end{tabular}
}
\end{table*}
\subsection{Parameter Sensitivity Analysis}

We conduct a sensitivity analysis to evaluate the robustness of SF-GNN under different hyperparameter settings. Table~\ref{tab:sensitivity} examines the effects of the edge dropping probability ($p$), hub quantile ($q$), downweight factor ($\gamma$), and fairness regularization strength ($\lambda$) across five benchmark datasets: NBA, German, Credit, Income, and Pokec\_n. For each parameter group, we report results under low, medium, and high settings while fixing all remaining hyperparameters.

Overall, the results demonstrate that SF-GNN remains relatively stable under moderate parameter variations, while still exhibiting meaningful fairness--utility trade-offs. Increasing fairness-related regularization generally improves $\Delta$SP and $\Delta$EO at the cost of small reductions in predictive accuracy. For example, lower $\lambda$ values achieve the best fairness performance on NBA and Pokec\_n, while larger $\lambda$ values improve accuracy on Credit. Similarly, moderate edge filtering and downweighting settings typically provide the best balance between fairness and predictive utility, whereas overly aggressive filtering can degrade performance by removing useful structural information. Across datasets, the selected SF-GNN configuration consistently achieves competitive or best overall trade-offs between ACC, $\Delta$SP, and $\Delta$EO, indicating that the proposed framework is robust and not overly sensitive to precise hyperparameter tuning.
\begin{table*}[h!]
\centering
\small
\caption{Sensitivity analysis of SF-GNN hyperparameters across datasets, evaluating the fairness regularization coefficient ($\lambda$), edge drop probability ($p$), downweight factor ($\gamma$), and hub quantile ($q$). Hyperparameters are tuned via grid search over $\lambda \in {0.5, 0.6, 0.7, 0.8}$, $p \in {0.1, 0.2, 0.3}$, $\gamma \in {0.7, 0.8, 0.9}$, and $q \in {0.4, 0.5, 0.6}$. For each parameter, Low/Mid/High correspond to the smallest, middle, and largest tested values. We report ACC ($\uparrow$), $\Delta$SP ($\downarrow$), and $\Delta$EO ($\downarrow$).}
\label{tab:sensitivity}
\resizebox{\linewidth}{!}{
\begin{tabular}{ll
|ccc
|ccc
|ccc
|ccc}
\toprule
\multirow{2}{*}{\textbf{Dataset}}
& \multirow{2}{*}{\textbf{Metric}}
& \multicolumn{3}{c|}{$p$}
& \multicolumn{3}{c|}{$q$}
& \multicolumn{3}{c|}{$\gamma$}
& \multicolumn{3}{c}{$\lambda$} \\
\cmidrule(lr){3-5}
\cmidrule(lr){6-8}
\cmidrule(lr){9-11}
\cmidrule(lr){12-14}
&
& Low & Mid & High
& Low & Mid & High
& Low & Mid & High
& Low & Mid & High \\
\midrule

\multirow{3}{*}{NBA}
& ACC$\uparrow$
& 71.90 & 70.13 & 71.65
& 71.14 & 71.65 & 71.90
& 71.14 & 70.38 & 71.66
& \textbf{72.15} & 70.89 & 71.39 \\

& $\Delta$SP$\downarrow$
& 4.86 & 5.03 & 8.45
& 6.89 & 7.16 & 5.32
& 5.15 & 7.46 & 4.84
& \textbf{2.29} & 6.44 & 9.25 \\

& $\Delta$EO$\downarrow$
& \textbf{7.66} & 8.11 & 18.68
& 19.07 & 12.91 & 9.66
& 10.49 & 10.10 & 11.62
& 9.10 & 8.54 & 8.76 \\
\midrule

\multirow{3}{*}{German}
& ACC$\uparrow$
& 70.40 & 70.40 & 70.16
& 70.00 & 69.44 & 70.40
& 70.00 & 70.16 & 70.40
& \textbf{70.56} & 70.16 & 69.68 \\

& $\Delta$SP$\downarrow$
& \textbf{0.63} & 1.99 & 2.90
& 1.57 & 1.68 & 1.53
& 2.06 & 1.66 & 2.27
& 3.12 & 1.42 & 3.52 \\

& $\Delta$EO$\downarrow$
& 2.92 & 4.59 & 3.21
& 3.44 & 4.42 & 3.09
& \textbf{2.59} & 4.14 & 3.79
& 6.44 & 5.70 & 4.56 \\
\midrule

\multirow{3}{*}{Credit}
& ACC$\uparrow$
& 72.27 & 72.31 & 72.30
& 72.47 & 72.31 & 72.24
& 72.31 & 72.31 & 72.31
& 72.11 & 72.31 & \textbf{72.51} \\

& $\Delta$SP$\downarrow$
& 1.08 & 1.19 & 1.25
& 1.33 & 1.18 & \textbf{0.92}
& 1.15 & 1.19 & 1.11
& 1.02 & 1.18 & 1.48 \\

& $\Delta$EO$\downarrow$
& 3.02 & 3.05 & 2.95
& 3.29 & 3.05 & 3.39
& 3.24 & 3.05 & 3.29
& 3.38 & 3.05 & \textbf{2.81} \\
\midrule

\multirow{3}{*}{Income}
& ACC$\uparrow$
& \textbf{74.58} & 73.88 & 74.12
& 74.20 & 74.03 & 74.04
& 74.06 & 73.96 & 74.02
& 74.14 & 73.87 & 73.85 \\

& $\Delta$SP$\downarrow$
& 3.38 & 2.81 & 3.25
& 2.98 & 3.02 & 3.19
& 2.90 & 3.14 & 3.26
& 2.87 & 3.40 & \textbf{2.60} \\

& $\Delta$EO$\downarrow$
& 3.06 & 1.88 & 2.76
& 3.39 & 1.94 & 2.20
& 2.36 & 2.21 & 1.94
& 2.62 & \textbf{1.47} & 2.39 \\
\midrule

\multirow{3}{*}{Pokec\_n}
& ACC$\uparrow$
& \textbf{91.31} & 91.18 & 91.19
& 91.26 & 91.26 & 91.21
& 91.26 & 91.25 & 91.20
& 91.30 & 91.19 & 91.22 \\

& $\Delta$SP$\downarrow$
& 1.11 & 1.13 & 1.12
& 1.23 & 1.11 & 1.09
& 1.18 & 1.15 & 1.17
& \textbf{1.06} & 1.17 & 1.16 \\

& $\Delta$EO$\downarrow$
& \textbf{3.26} & 4.24 & 4.13
& 4.48 & 4.50 & 5.43
& 5.58 & 4.21 & 4.51
& 3.72 & 4.53 & 4.51 \\
\bottomrule
\end{tabular}
}
\end{table*}

\section{Conclusion}
This paper studied fairness in GNNs from a structural
perspective, motivated by the observation that sensitive-group correlations
embedded in graph topology can be repeatedly reinforced through message
passing. To mitigate this topology-induced bias, we proposed SF-GNN, a simple
and architecture-agnostic framework that identifies bias-prone edges by
combining sensitive homophily with structural propagation amplifiers and selectively attenuates these routes via stochastic edge dropping and downweighting during aggregation. SF-GNN
is lightweight, requires only a one-time preprocessing step, and optimizes a
composite objective with a warm-up statistical-parity regularizer for stable
training. 

Across five benchmark datasets, SF-GNN achieves a favorable
fairness--accuracy trade-off compared with standard GNNs, graph transformer
models, and recent fairness-aware baselines, consistently reducing statistical
parity disparity while maintaining competitive predictive performance.
Ablation studies further confirm that the gains arise from targeted structural
filtering and the joint design of bias-prone edge identification and training
dynamics, rather than from modifying a large number of edges.

Future work includes extending SF-GNN to additional fairness notions (e.g.,
equalized odds), multi-class and multi-attribute settings, and learning
adaptive filtering policies that account for heterophily and temporal graphs.
More broadly, these results suggest that explicitly regulating bias propagation
paths provides an effective and practical direction for fairness-aware graph
learning in high-stakes applications.

\subsubsection{\discintname}
The authors declare that they have no competing interests relevant to the content of this article.

%
%
%
\bibliographystyle{splncs04}
\bibliography{mybibliography}

@inproceedings{dai2021say,
  title={Say no to the discrimination: Learning fair graph neural networks with limited sensitive attribute information},
  author={Dai, Enyan and Wang, Suhang},
  booktitle={Proceedings of the 14th ACM international conference on web search and data mining},
  pages={680--688},
  year={2021}
}

@inproceedings{agarwal2021towards,
  title={Towards a unified framework for fair and stable graph representation learning},
  author={Agarwal, Chirag and Lakkaraju, Himabindu and Zitnik, Marinka},
  booktitle={Uncertainty in artificial intelligence},
  pages={2114--2124},
  year={2021},
  organization={PMLR}
}

@inproceedings{ling2023learning,
  title={Learning fair graph representations via automated data augmentations},
  author={Ling, Hongyi and Jiang, Zhimeng and Luo, Youzhi and Ji, Shuiwang and Zou, Na},
  booktitle={International Conference on Learning Representations (ICLR)},
  year={2023}
}

@inproceedings{ijcai2024p50,
  title     = {FairGT: A Fairness-aware Graph Transformer},
  author    = {Luo, Renqiang and Huang, Huafei and Yu, Shuo and Zhang, Xiuzhen and Xia, Feng},
  booktitle = {Proceedings of the Thirty-Third International Joint Conference on
               Artificial Intelligence, {IJCAI-24}},
  publisher = {International Joint Conferences on Artificial Intelligence Organization},
  editor    = {Kate Larson},
  pages     = {449--457},
  year      = {2024},
  month     = {8},
  note      = {Main Track},
}

@inproceedings{dong2023interpreting,
  title={Interpreting unfairness in graph neural networks via training node attribution},
  author={Dong, Yushun and Wang, Song and Ma, Jing and Liu, Ninghao and Li, Jundong},
  booktitle={Proceedings of the aaai conference on artificial intelligence},
  volume={37},
  number={6},
  pages={7441--7449},
  year={2023}
}

@article{kipf2016semi,
  title={Semi-supervised classification with graph convolutional networks},
  author={Kipf, Thomas N and Welling, Max},
  journal={arXiv preprint arXiv:1609.02907},
  year={2016}
}

@misc{asuncion2007uci,
  title={UCI machine learning repository},
  author={Asuncion, Arthur and Newman, David and others},
  year={2007},
  publisher={Irvine, CA, USA}
}

@article{yeh2009comparisons,
  title={The comparisons of data mining techniques for the predictive accuracy of probability of default of credit card clients},
  author={Yeh, I-Cheng and Lien, Che-hui},
  journal={Expert systems with applications},
  volume={36},
  number={2},
  pages={2473--2480},
  year={2009},
  publisher={Elsevier}
}

@article{fey2019fast,
  title={Fast graph representation learning with PyTorch Geometric},
  author={Fey, Matthias and Lenssen, Jan Eric},
  journal={arXiv preprint arXiv:1903.02428},
  year={2019}
}

@article{velivckovic2017graph,
  title={Graph attention networks},
  author={Veli{\v{c}}kovi{\'c}, Petar and Cucurull, Guillem and Casanova, Arantxa and Romero, Adriana and Lio, Pietro and Bengio, Yoshua},
  journal={arXiv preprint arXiv:1710.10903},
  year={2017}
}

@article{dwivedi2020generalization,
  title={A generalization of transformer networks to graphs},
  author={Dwivedi, Vijay Prakash and Bresson, Xavier},
  journal={arXiv preprint arXiv:2012.09699},
  year={2020}
}

@article{dai2022learning,
  title={Learning fair graph neural networks with limited and private sensitive attribute information},
  author={Dai, Enyan and Wang, Suhang},
  journal={IEEE Transactions on Knowledge and Data Engineering},
  volume={35},
  number={7},
  pages={7103--7117},
  year={2022},
  publisher={IEEE}
}

@article{zhou2020graph,
  title={Graph neural networks: A review of methods and applications},
  author={Zhou, Jie and Cui, Ganqu and Hu, Shengding and Zhang, Zhengyan and Yang, Cheng and Liu, Zhiyuan and Wang, Lifeng and Li, Changcheng and Sun, Maosong},
  journal={AI open},
  volume={1},
  pages={57--81},
  year={2020},
  publisher={Elsevier}
}

@article{yang2026relational,
  title={Relational Graph Modeling for Credit Default Prediction: Heterogeneous GNNs and Hybrid Ensemble Learning},
  author={Yang, Yvonne and Vasistha, Eranki},
  journal={arXiv preprint arXiv:2601.14633},
  year={2026}
}

@inproceedings{purificato2025gnn,
  title={GNN’s FAME: Fairness-Aware MEssages for Graph Neural Networks},
  author={Purificato, Erasmo and Mahadik, Hannan Javed and Boratto, Ludovico and De Luca, Ernesto William},
  booktitle={Proceedings of the 33rd ACM Conference on User Modeling, Adaptation and Personalization},
  pages={301--306},
  year={2025}
}

@article{wang2024toward,
  title={Toward fair graph neural networks via real counterfactual samples},
  author={Wang, Zichong and Qiu, Meikang and Chen, Min and Salem, Malek Ben and Yao, Xin and Zhang, Wenbin},
  journal={Knowledge and Information Systems},
  volume={66},
  number={11},
  pages={6617--6641},
  year={2024},
  publisher={Springer}
}

@inproceedings{sium2024individual,
  title={Individual Fairness in Graphs Using Local and Global Structural Information},
  author={Sium, Yonas and Li, Qi and Varshney, Kush R},
  booktitle={Proceedings of the AAAI/ACM Conference on AI, Ethics, and Society},
  volume={7},
  number={1},
  pages={1379--1389},
  year={2024}
}

@article{zhu2024fairagg,
  title={FairAGG: Toward fair graph neural networks via fair aggregation},
  author={Zhu, Yuchang and Li, Jintang and Chen, Liang and Zheng, Zibin},
  journal={IEEE Transactions on Computational Social Systems},
  volume={11},
  number={5},
  pages={6308--6319},
  year={2024},
  publisher={IEEE}
}

@inproceedings{yang2024fairsin,
  title={Fairsin: Achieving fairness in graph neural networks through sensitive information neutralization},
  author={Yang, Cheng and Liu, Jixi and Yan, Yunhe and Shi, Chuan},
  booktitle={Proceedings of the AAAI conference on artificial intelligence},
  volume={38},
  number={8},
  pages={9241--9249},
  year={2024}
}

@article{jiang2022fmp,
  title={Fmp: Toward fair graph message passing against topology bias},
  author={Jiang, Zhimeng and Han, Xiaotian and Fan, Chao and Liu, Zirui and Zou, Na and Mostafavi, Ali and Hu, Xia},
  journal={arXiv preprint arXiv:2202.04187},
  year={2022}
}

@inproceedings{takac2012data,
  title={Data analysis in public social networks},
  author={Takac, Lubos and Zabovsky, Michal},
  booktitle={International scientific conference and international workshop present day trends of innovations},
  volume={1},
  number={6},
  year={2012}
}

@article{wu2021representing,
  title={Representing long-range context for graph neural networks with global attention},
  author={Wu, Zhanghao and Jain, Paras and Wright, Matthew and Mirhoseini, Azalia and Gonzalez, Joseph E and Stoica, Ion},
  journal={Advances in neural information processing systems},
  volume={34},
  pages={13266--13279},
  year={2021}
}

\end{document}